\documentclass{llncs}
\usepackage{amsmath, amssymb,bm}
\usepackage{xcolor}
\usepackage{booktabs}
\usepackage{graphicx}
\usepackage{color}
\usepackage{multirow}
\usepackage{algorithm}  
\usepackage{tabularx}
\usepackage{amsfonts}
\usepackage{algpseudocode}  
\usepackage{multirow}
\usepackage{amsmath,booktabs}
\usepackage{algpseudocode}  
\usepackage{multirow}
\usepackage{amsmath,booktabs}

\begin{document}

\title{Deeply Supervised Rotation Equivariant 
	Network for Lesion Segmentation in Dermoscopy Images}
\author{Xiaomeng Li, Lequan Yu, Chi-Wing Fu and Pheng-Ann Heng}
\institute{Department of Computer Science and Engineering, The Chinese University of Hong Kong, Hong Kong}

\maketitle
\begin{abstract}
Automatic lesion segmentation in dermoscopy images is an essential step for computer-aided diagnosis of melanoma.
The dermoscopy images exhibits rotational and reflectional symmetry, however, this geometric property has not been encoded in the state-of-the-art convolutional neural networks based skin lesion segmentation methods.   
In this paper, we present a deeply supervised rotation equivariant network for skin lesion segmentation by extending the recent group rotation equivariant network~\cite{cohen2016group}.   
Specifically, we propose the G-upsampling and G-projection operations to adapt the rotation equivariant classification network for our skin lesion segmentation problem.
To further increase the performance, we integrate the deep supervision scheme into our proposed rotation equivariant segmentation architecture.
The whole framework is equivariant to input transformations, including rotation and reflection, which improves the network efficiency and thus contributes to the segmentation performance.
We extensively evaluate our method on the ISIC 2017 skin lesion challenge dataset.
The experimental results show that our rotation equivariant networks consistently excel the regular counterparts with the same model complexity under different experimental settings.
Our best model also outperforms the state-of-the-art challenging methods, which further demonstrate the effectiveness of our proposed deeply supervised rotation equivariant segmentation network.
\end{abstract}
\section{Introduction}
Skin cancer has become the most prevalent cancer in the United States~\cite{rogers2015incidence}, and melanoma is the most deadly form of skin cancer, leading to over 9,000 deaths in the Unite States in 2017~\cite{CAAC:CAAC21387}.
A common technique used by dermatologists for diagnosing skin diseases is the dermoscopy, which enables observation by enhancing the visual effect of pigmented skin lesions.
Lesion segmentation in dermoscopy images is an essential component in the diagnosis of skin diseases.
However, segmenting skin lesions by dermatologists is time-consuming and error-prone to inter- and intra-observer variabilities. 
Moreover, due to the growing shortage of dermatologists per capita, the automatic lesion segmentation in dermoscopy images would be beneficial to more people~\cite{kimball2008us}.
Convolutional neural networks (CNNs) have proven to be very powerful models for a board array of image recognition tasks. 
In the domain of skin lesion segmentation, all leading methods adopted CNN-based methods~\cite{yuan2017improving,berseth2017isic,yu2017automated}.
For example, Yuan et al.~\cite{yuan2017improving} proposed a deep convolutional neural network (DCNN), trained it with multiple color spaces, and achieved the best performance in the ISIC 2017 skin lesion segmentation challenge.
Yu et al.~\cite{yu2017automated} explored the network depth property and proposed a deep residual network with more than 50 layers for automatic skin lesion segmentation. 

\begin{figure}[!t]
	\centering
	\includegraphics[width=0.85\linewidth]{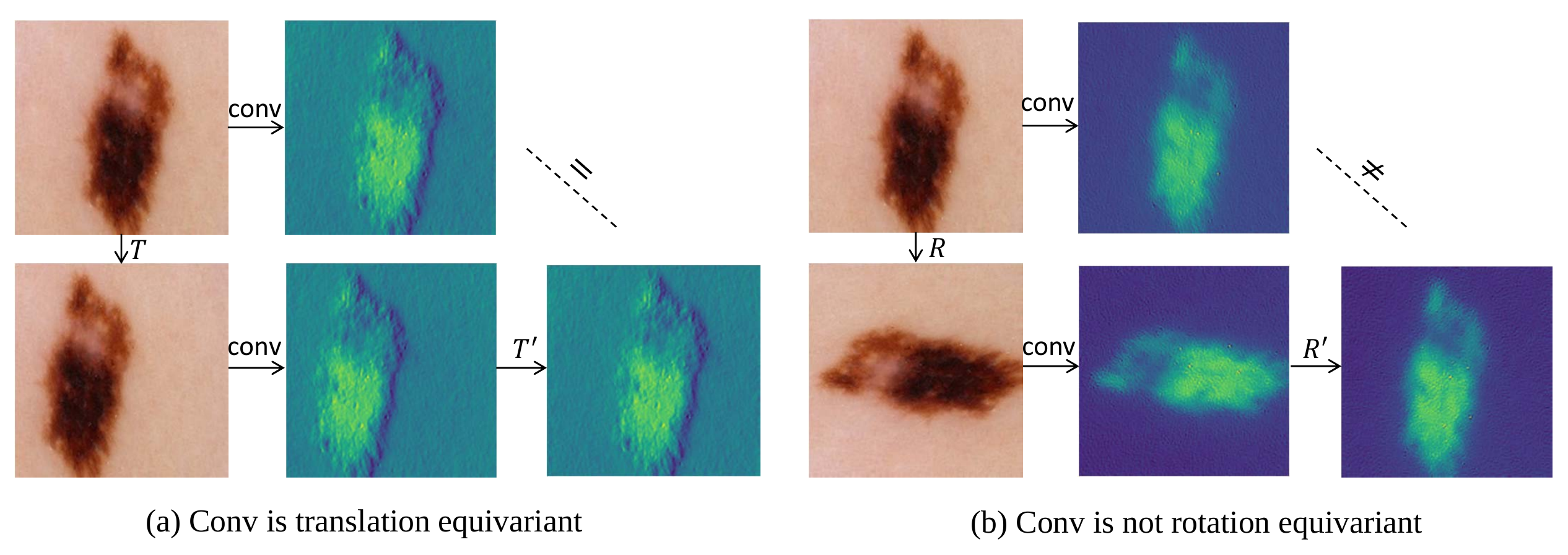}
	\caption{Convolution layer is translation equivariant (a); but convolution is not rotation equivariant (Zoom in to see the detailed comparison), as shown in (b).}
	\label{fig:intro}\centering
	\vspace{-0.3cm}
\end{figure}

The success of these CNN-based models can be partially attributed to the effectiveness of weights sharing in the convolution layer, where the translation equivariance is preserved. 
To be specific, translating a layer's input produces the corresponding translation in the layer's output. As shown in Fig.~\ref{fig:intro}(a), shifting the input of the convolution leads to the predictable shifting in the output.
This translation equivariance property of convolution is effective in most perception tasks, where the same weights can be used to encode the local spatial pattern and reduce the model parameter to avoid overfitting. 
Unlike natural images, dermoscopy images exhibit not only translation symmetry but also rotation and flipping symmetry as well.
However, if one rotates the convolution input, the generated output does not necessarily rotate in a predictable manner, as shown in Fig.~\ref{fig:intro}(b). 
Previous works utilized data augmentation technique like rotation and flipping, to encourage the network to learn rotation and flipping covariance. 
Even though this strategy could regularize the network to learn the equivariance on the training set, there is no guarantee that the equivariance property will generalize to other images.
Moreover, forcing the network to learn the redundant knowledge introduced by different data transformations would reduce the model efficiency.
Specifically, with the same level of model complexity, the regular CNN needs to learns not only the discriminative features but only the input rotations and reflections.
Furthermore, comparing with natural images, the biomedical images are scarce and more difficult to obtain, and it is highly demanded to design an efficient network to improve the model efficiency.

We consider to improve the network efficiency by encoding the rotation and flipping equivariance into the network, in which the network preserves the equivariance inherent without relying on data augmentation. 
Recently, there are some works have made significant progress for rotation equivariant networks~\cite{cohen2016group,marcos2017rotation}. 
Cohen \emph{et al.}~\cite{cohen2016group} explored rotation and reflection equivariant inherent network for classification problems, where the feature learned in the $G$ space exhibits rotation equivariance.
In this paper, we propose a deeply supervised rotation equivariant network by extending G-CNN~\cite{cohen2016group} for skin lesion segmentation.
Our network encodes the translation, rotation and flipping symmetry of dermoscopy images, and thus improves the skin lesion segmentation performance.
Specifically, we design the $G$-upsampling layer and the $G$-projection layer for the segmentation task with the $G$-convolution layer.
The $G$-upsampling layer upsamples the features in the $G$ space and the $G$-projection layer performs average pooling over the rotation dimension and then projects features from the $G$ space to $\mathbb{Z}$ space, making the whole network rotation equivariant.
To better stabilize the learning processing of the proposed network, we also integrated the deep supervision~\cite{lee2015deeply,chen2016dcan} in our network to further improve the performance.
Compared with the plain convolution neural networks, our network enjoys a substantially higher degree of weight sharing, and increases the expressive capacity of the network without increasing the number of parameters. 
We extensively evaluate our method on the ISIC 2017 skin lesion segmentation challenge. The results demonstrate the efficiency of our proposed rotation equivariant segmentation network, and our method outperforms other state-of-the-art methods on the challenging dataset.
Several works~\cite{bekkers2018roto,winkens2018improved,veeling2018rotation} also explore the rotation equivariant network in the biomedical image domain.
However, our work further explores the equivariant segmentation networks with deep supervision scheme~\cite{lee2015deeply,chen2016dcan} for automatic lesion segmentation in dermoscopy images.

\section{Method}
In this section, we first introduce the concept of group equivariant convolution ($G$-convolution), and then describe the proposed $G$-upsampling and $G$-projection layers for the segmentation task. Finally we present our proposed deeply supervised rotation equivariant framework. 

\subsection{$G$-convolution}

\begin{figure}[!t]
	\centering
	\includegraphics[width=0.7\linewidth]{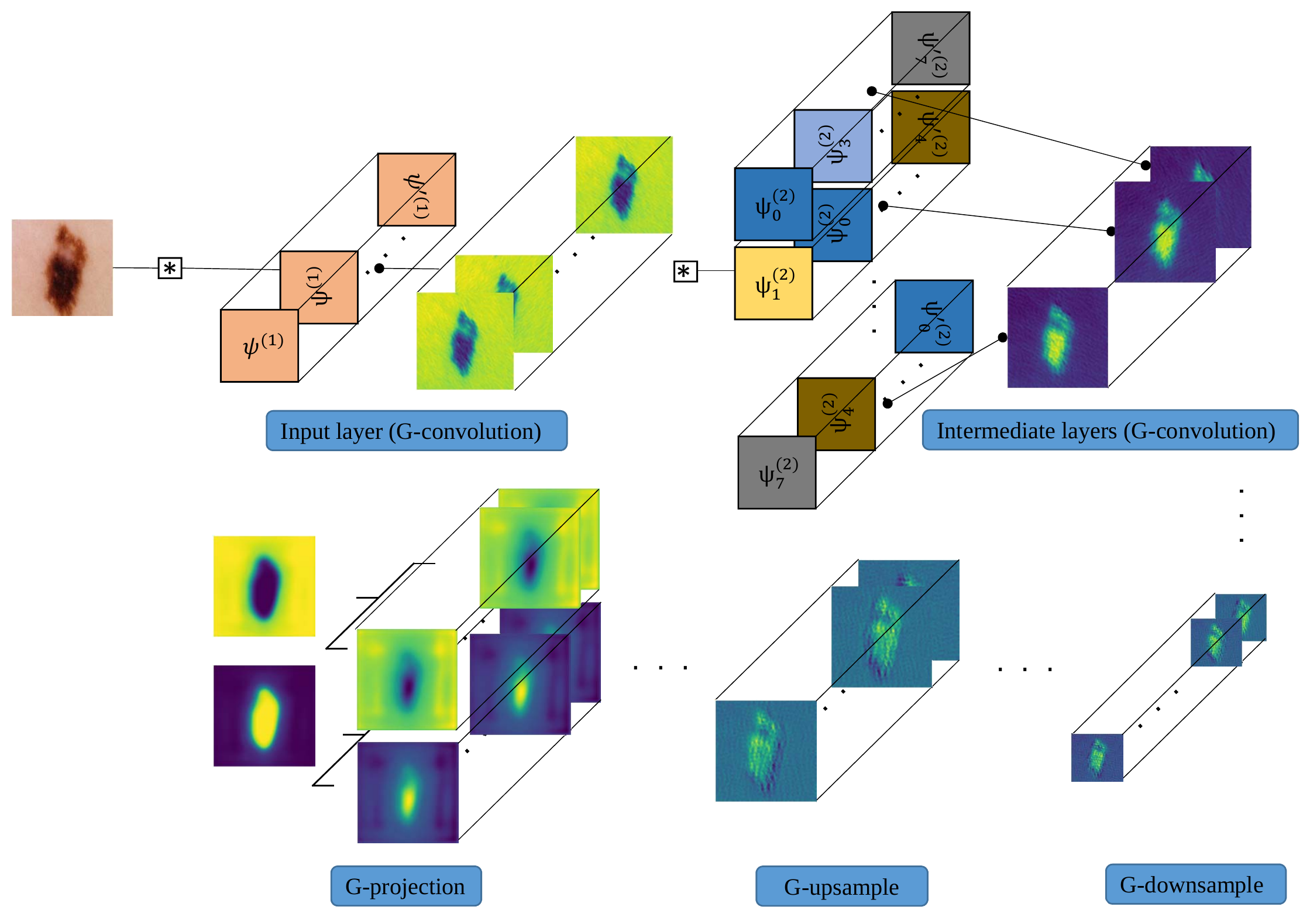}
	\caption{The illustration of the $G$-convolution, $G$-upsampling and $G$-projection operation. Except the $G$-projection layer, we only show 1 channel in all other layers to simplify the illustration.}
	\label{fig:method}\centering
	\vspace{-0.3cm}
\end{figure} 
The regular first convolution layer is a function that maps the input to feature maps with $K$ channels $f: \mathbb{Z}^2 \rightarrow \mathbb{R}^K$. The function can be described as Eq.~\ref{eq:1}. 
\begin{equation}
[f * \varphi](x) = \sum _{y \in \mathbb{Z}^{2}}  \sum_k f_k(y) \varphi _k (x-y),
\label{eq:1}
\end{equation}
where $\varphi_k$ denotes the convolution kernel.

To encode rotation equivariance in the network, Cohen \emph{et al.}~\cite{cohen2016group} proposed to conduct convolution on groups, where the group $p4$ consists of all compositions of translations and rotations by $90^\circ$ about any center of rotation in the grid, and the group $p4m$ additionally includes reflections. 
Specifically, for the input layer, the ($\mathbb{Z}^2 \rightarrow G$) convolution is defined as
\begin{equation}
[f * \varphi](g) = \sum _{y \in \mathbb{Z}^{2}}  \sum_k f_k(y) \varphi _k (g^{-1}y),
\end{equation}
where $g$ is a transformation in the predefined group $p4$ or $p4m$.
Then, in the following layers, feature maps and filters are both functions on $G$ and the ($ G \rightarrow G $) convolution can be described as
\begin{equation}
[f * \varphi](g) = \sum _{h \in G}  \sum_k f_k(h) \varphi _k (g^{-1}h)
\end{equation}

\subsection{$G$-upsampling and $G$-projection for Segmentation Problem}
In the segmentation problem, the down-sampled feature maps need to be upsampled in the $G$ space for pixel-level prediction, and thus we design the $G$-upsampling layer. The convention upsampling layer performs upsample operation for feature maps at the spatial dimension. In the $G$ space, the $G$-upsampling layer performs upsample operation over all eight rotations (for group $p4m$) at each spatial position, as shown in Fig.~\ref{fig:method}. 

To enable the equivariant network to produce final score maps for skin lesion segmentation, we also define the ($G-\mathbb{Z}^2$) projection layer. 
\begin{equation}
f_k(y) = \frac {1}{|G|} { \sum_{G} \\ (f_k(h))},
\end{equation}
where $|G|$ denotes the number of element in group $G$. For example, it equals to 4 for group $p4$ and 8 for group $p4m$. With the $G$-upsampling layer and the $G$-projection layer, we can design a segmentation network, which is equivariant to the input symmetric transformations.

\begin{figure}[!t] 
	\centering
	\includegraphics[width=0.85\linewidth]{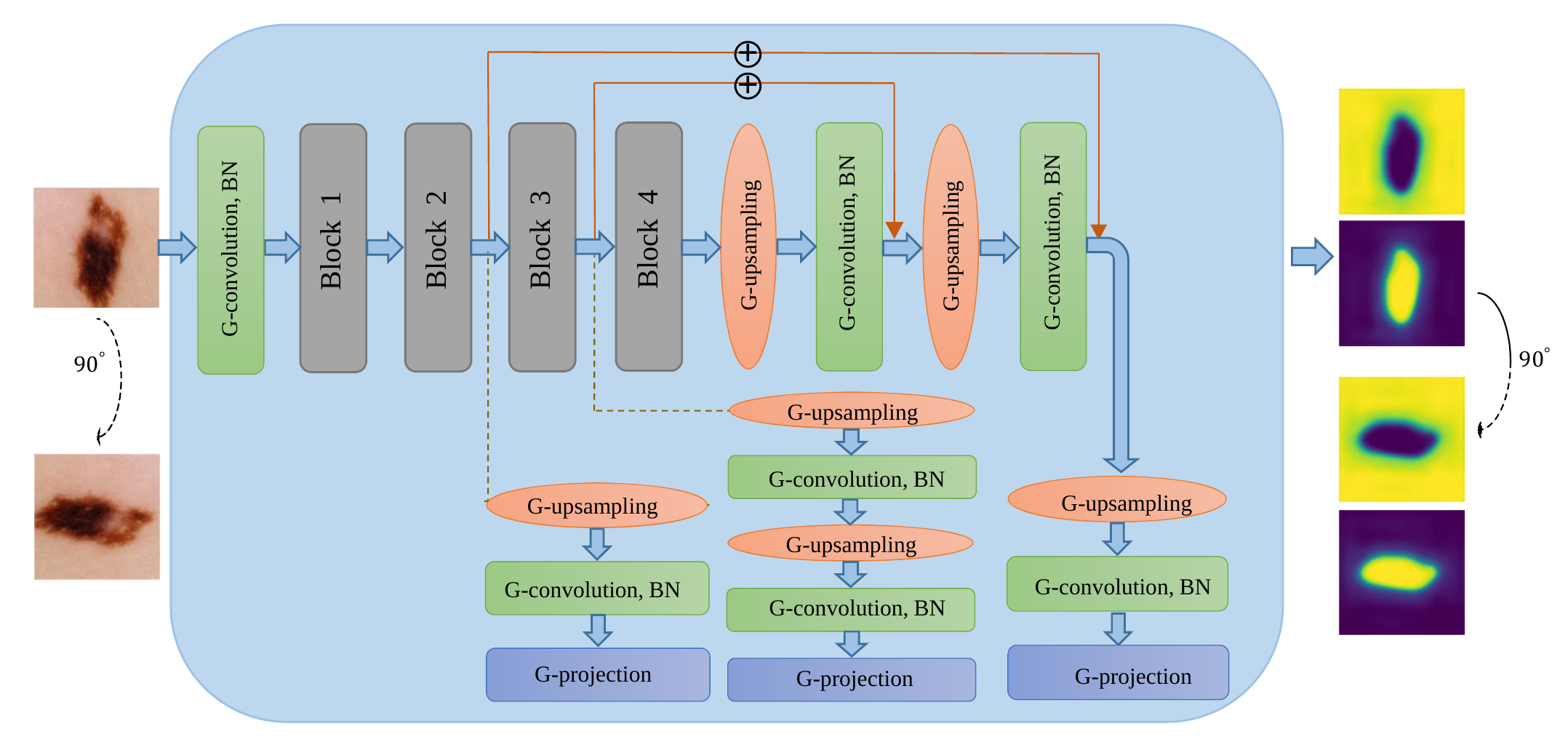}
	\caption{The framework of our proposed rotation equivariant network for skin lesion segmentation. The network is based on ResNet34 backbone, and is integrated with deep supervision and U-Net connections. All the regular operations are replaced to G-convolution, G-upsampling, and G-projection operations.
		The whole architecture is equivariant to input symmetric transformation. 
		In other words, if one rotate the input for $90^\circ$, then the prediction score would rotate in the same way. Note that we omit the pooling operation, ReLU activations to simplify the illustration.}
	\label{fig:framework}\centering
	\vspace{-0.3cm}
\end{figure} 

\subsection{Deeply Supervised $G$-FCNs}

The deeply supervised rotation equivariant network is based on the ResNet34~\cite{he2016deep} architecture, where we replace the convolution layer, upsampling layer to the G-convolution, G-upsampling and G-projection layers.
As shown in Fig.~\ref{fig:framework}, we use three $2\times2$ G-upsampling layers and one G-projection layer following the feature maps generated by ResNet34.
We also adopt the U-net like long-skip connections to preserve the low-level features.
The deep supervision mechanism is performed by upsampling at three different spatial resolution of features, and the final result is the weighted combination of three segmentation predictions. 
Since all the elements in the network are equivariant to $90^ \circ$ rotation and reflection of the input, the whole framework also preserves the rotation equivariant property.
In other words, if one clock-wise rotates the input image $90^{\circ}$, the network output will rotate in the same manner. 
Our code will be made available and readers can find more details about the network architecture\footnote{https://github.com/xmengli999/Deeply-Supervised-Rotation-Equivariant-Network-for-Lesion-Segmentation}.
\section{Experiments and Results}

\subsection{Dataset and Evaluation Metrics}
We evaluate our method on the dataset of ISIC 2017 skin lesion segmentation challenge~\cite{codella2017skin}, which consists of a training set with 2000 annotated dermoscopic images, a validation set with 150 images, and a testing set with 600 images.
The image size ranges from $540 \times 722$ to $4499 \times 6748$. To keep balance between segmentation performance and computational cost, we first resize all the images to $224 \times 224$ using bicubic interpolation. 
For evaluation metric, we follow the challenge instructions to employ five evaluation metrics, including jaccard index (JA), dice coefficient (DI), pixel-wise accuracy (AC), sensitivity (SE) and specificity (SP). 
Note that the final rank is determined according to JA in the ISIC 2017 skin lesion segmentation challenge.

\subsection{Implementation Details}
All the experiments were implemented using PyTorch~\cite{paszke2017automatic}, and were trained with stochastic gradient descent (SGD) algorithm (momentum is 0.9) from scratch.
The learning rate is set to 0.01 and decays at epoch 60. All the models are trained for 70 epochs.
As for experiments with the plain convolution, we employed data augmentation like $90^\circ$ rotation and flipping. 
The weights for deep supervision are 0.7, 0.2 and 0.1 respectively.

\subsection{Ablation Study}


\begin{table}[t]
	\centering
	\caption{Ablation study of the deeply supervised rotation equivariant network.}
	\label{tab:50labels} %
	{	
		\begin{tabular}{c|c|c|c|c|c|c}
			\hline
			\multirow{2}{*}{Model} & \multirow{2}*{No. of Para} & \multicolumn{5}{c}{Evaluation Metrics} 
			\tabularnewline
			\cline{3-7} &
			& JA & DI
			& AC & SE & SP \tabularnewline
			\hline
			ResnetFCN34* & 22.8M & 71.27 & 80.21  & 91.39  & 78.31 & 96.78 \tabularnewline
			\hline
			(RE)-ResnetFCN34* & 22.8M  & 74.54 & 83.27 &  92.58  & 81.05 & \textbf{97.59}  \tabularnewline 
			\hline
			DS-U-ResnetFCN34* & 23.2M & 74.38 & 83.06 & 92.51 & 82.52 & 97.14 \tabularnewline
			\hline
			\hline
			(RE)-DS-U-ResnetFCN34* (ours) & 23.2M & 76.65 & 85.00 &  93.27  & 84.61 & 96.80  \tabularnewline 
			\hline
			
			
			(RE)-DS-U-ResnetFCN34 (ours) & 23.5M & \textbf{77.23} & \textbf{85.60} &  \textbf{93.55}  & \textbf{85.40} & 97.15 \tabularnewline
			\hline
		\end{tabular}
	}
\end{table}
Table~\ref{tab:50labels} shows the segmentation performance on the test dataset with different configurations.
ResnetFCN34* refers to the FCN-based Resnet34 network, while (RE)-ResnetFCN34* and DS-U-ResnetFCN34* are the rotation equivariant and deeply supervised with long range U-Net connections counterparts, respectively.
The * denotes that we remove the first pooling layer from the original Resnet34 network, following the setting in \cite{cohen2016group}. 
Note that all the rotation equivariant networks are performed with group $p4m$~\cite{cohen2016group}. 
To analyze the effectiveness of rotation equivariant network fairly, all the comparison are performed with the same model complexity. Specifically, compared with the original filter numbers in Resnet34, the number of filters is divided by roughly $\sqrt{8}$ in each G-convolution layer.

From the comparison in Table~\ref{tab:50labels}, we can see that the rotation equivariant network largely excels the plain counterpart, with 3.27\% improvement on JA. 
The deeply supervised version also improve the JA performance significantly.
When integrate the deep supervision with U-Net connections into the rotation equivariant network ((RE)-DS-U-ResnetFCN34*), we can further improve the segmentation performance (2.27\% on JA).
To better adapt the network for our skin lesion segmentation task, we replace the first pooling layer of ResnetFCN34 with a G-convolution with stride of 2 and denoted the deeply supervised rotation equivariant version as (RE)-DS-U-ResnetFCN34.
It is observed that (RE)-DS-U-ResnetFCN34 achieves the best performance on the all evaluation metrics excepting for SP, demonstrating the superiority and effectiveness of rotation equivariant networks under same level of model complexity.

\subsection{Comparison with Other Methods}
\begin{table*}[!t]
	\centering
	\caption{Comparison with state-of-the-art methods on the ISIC 2017 test dataset.}
	\label{tab: testdata} %
	{
		\begin{tabular}{c|c|c|c|c|c}
			\toprule[1.5pt]
			Team & JA  & DI  & AC & SE & SP \tabularnewline  \hline \hline
			Our  Method
			&\textbf{0.772} & \textbf{0.856} &  \textbf{0.936}  & \textbf{0.854} & 0.972
			\tabularnewline
			\hline
			Yuan and Lo~\cite{yuan2017improving} & 0.765 & 0.849   & 0.934  & 0.825 & 0.975  \tabularnewline
			\hline
			Berseth~\cite{berseth2017isic} & 0.762 & 0.847 &  0.932 & 0.820 & 0.978  \tabularnewline
			\hline
			Bi et al.~\cite{bi2017automatic} & 0.760 & 0.844 &  0.934  & 0.802 & \textbf{0.985}  \tabularnewline
			\hline
			RECOD & 0.754 & 0.839 & 0.931 & 0.817 & 0.970  \tabularnewline
			\hline		
			Jer & 0.752 & 0.837 &  0.930 & 0.813 & 0.976    	  
			\tabularnewline
			\hline			
			NedMos & 0.749 & 0.839 &  0.930 & 0.810 & 0.981
			\tabularnewline
			\hline
			INESC & 0.735 & 0.824 &  0.922 & 0.813 & 0.968
			\tabularnewline
			\hline
			Shenzhen U (Lee) & 0.718 & 0.810 &  0.922 & 0.789 & 0.975  \tabularnewline
			\bottomrule[1.5pt]	  
		\end{tabular}}
	\end{table*}
We compare our result with state-of-the-art results on the ISIC 2017 testing dataset. There are totally 21 submissions and the top results are listed in Table 2.
Yuan \emph{et al.}~\cite{yuan2017improving} trained a CNN network with multiple color spaces and achieves the 
best performance on the skin lesion segmentation challenge. 
Our best model, trained from scratch on the single RGB color space, outperforms other state-of-the-arts in the test dataset of the ISIC challenge. This comparison validates the effectiveness of our proposed deeply supervised rotation equivariant network in the skin lesion segmentation task. 

\section{Conclusion}
In this paper, we present a deeply supervised rotation equivariant segmentation network for skin lesion segmentation by utilizing the recent findings on rotation equivariant CNNs. 
We design the G-upsampling and G-projection layers to enable our network for the segmentation task, and introduce the deep supervision mechanism to improve performance.
Our network encodes the rotation and reflection symmetry of dermoscopy images, and significantly improves the skin lesion segmentation performance.
Our method has achieved the best performance on the ISIC 2017 skin lesion segmentation challenge dataset.
Future works include the extension of equivariance to arbitrary rotation and scaling.   

\section{Acknowledgment}
We special thank Dr. Taco Cohen for fruitful discussions about their work, kindly help and encouragement in our exploration.

\bibliographystyle{splncs04}
\bibliography{refs}
\end{document}